\documentclass{article}
\usepackage{algorithm}
\usepackage{algpseudocode}
\PassOptionsToPackage{numbers,sort,compress}{natbib}
\usepackage{wrapfig}
\usepackage{float}
\usepackage{enumitem}

\usepackage[final]{neurips_mlsb_2025}

\usepackage[utf8]{inputenc} 
\usepackage[T1]{fontenc}    
\usepackage[hidelinks]{hyperref}       
\usepackage{url}            
\usepackage{booktabs}       
\usepackage{amsfonts}       
\usepackage{nicefrac}       
\usepackage{microtype}      
\usepackage{xcolor}         
\usepackage{amsmath, amssymb}
\usepackage{graphicx}
\usepackage{subcaption}

\title{g-DPO: Scalable Preference Optimization for Protein Language Models}

\author{%
\textbf{Constance Ferragu} \quad
\textbf{Jonathan D. Ziegler} \quad
\textbf{Nicolas Deutschmann} \quad \\
\textbf{Arthur Lindoulsi} \quad
\textbf{Eli Bixby} \quad
\textbf{Cradle ML Team} \\
Cradle \\
Zürich, Switzerland \\
\texttt{constance@cradle.bio} \\
}

\begin{document}

\maketitle

\begin{abstract}
Direct Preference Optimization (DPO) is an effective approach for aligning protein language models with experimental design goals. However, DPO faces a scalability bottleneck: the number of possible training pairs grows quadratically with the number of labeled sequences, leading to prohibitive training times even for modestly sized datasets. We introduce g-DPO, a framework that (i) uses sequence space clustering to prune redundant pairs while preserving training signal, and (ii) amortizes likelihood computations with group-based approximations. Across three protein engineering tasks, g-DPO maintains \textit{in silico} and \textit{in vitro} performance that is statistically indistinguishable from standard DPO, while converging $1.7\times$ to $5.4\times$ times faster, with speedups that scale with dataset size and the structure of the underlying mutational landscape.
\end{abstract}

\section{Introduction}
Protein language models (PLMs), trained with self-supervised learning on large sequence datasets, capture relevant structure and function signals \cite{madani2023large, xiao2025protein, rao2021msa, lin2023evolutionary}. To use these models for protein engineering tasks, such as optimizing binding affinity, thermostability, or catalytic activity, they are typically fine-tuned on labeled experimental datasets, many of which naturally define preferences (e.g., higher fluorescence is better, lower toxicity is better).

Direct Preference Optimization (DPO) is a natural fit for this setting. DPO fine-tunes a language model using pairwise preferences, maximizing the likelihood of preferred sequences relative to dispreferred ones \cite{rafailovDirectPreferenceOptimization2023}, without requiring a separate reward model as in RLHF \cite{christiano2017deep}. Recent studies have applied DPO to protein sequence design \cite{azarGeneralTheoreticalParadigm2023, widatallaAligningProteinGenerative2024, stocco2025guidinggenerativeproteinlanguage, yang2025steering}, however, unlike in NLP, where explicit human preferences are available \cite{ouyang2022training, bai2022training}, experimental protein datasets often provide scalar labels. Constructing preferences by exhaustively comparing all samples is infeasible because the number of pairs grows quadratically with dataset size. Moreover, not all comparisons are equally informative. Distant comparisons in sequence space tend to collapse into coarse binary signals, while local comparisons capture subtle and often non-additive effects of a few mutations \cite{johnson2023epistasis, hayashi2006experimental}, providing a more valuable learning signal, which is crucial in late-stage optimization. 

We introduce group-DPO (g-DPO), a DPO framework for experimentally labeled protein data that addresses these challenges. This study contributes:  
\textbf{(1) Scalable preference sampling:} Sequence-space clustering prunes redundant comparisons and focuses training on informative local neighborhoods in input space.  
\textbf{(2) Efficient training:} Building on this clustering, we exploit union masking to amortize log-likelihood computations across groups of sequences, further improving convergence time.  
\textbf{(3) Empirical validation:} Across three protein mutational landscapes, g-DPO maintains the \textit{in silico} and \textit{in vitro} performance of standard DPO, while converging $1.7$ to $5.4$ times faster, with larger convergence time gains expected as the size of the dataset increases.

\section{Related work: Constructing pairs for DPO with PLMs}
\textbf{Direct Preference Optimization (DPO).} DPO fine-tunes language models directly from pairwise comparisons, avoiding the need for a separate reward model \cite{rafailovDirectPreferenceOptimization2023}. Given a distribution $\mathcal{D}$ of pairs of sequences $(y_w, y_l)$, where $y_w$ (winner) is preferred over $y_l$ (loser), a policy $\pi_\theta$ is optimized to increase the likelihood of sampling $y_w$ over $y_l$ under a regularization constraint formulated as a bound on the KL divergence between $\pi_\theta$ and a reference policy $\pi_\mathrm{ref}$. This objective is expressed as a loss function:
\begin{align}
\mathcal{L}_{\mathrm{DPO}}(\pi_\theta, \pi_{\mathrm{ref}}) =
- \mathbb{E}_{(y_w, y_l) \sim \mathcal{D}} \left[
\log \sigma \Big(
\beta \log \frac{\pi_\theta(y_w)}{\pi_{\mathrm{ref}}(y_w)}
- \beta \log \frac{\pi_\theta(y_l)}{\pi_{\mathrm{ref}}(y_l)}
\Big)
\right],
\end{align}

where $\beta>0$ is the regularization hyperparameter and $\sigma$ is the sigmoid function. A complete derivation is provided in ~\ref{dpoloss}. 
A primary challenge in applying DPO to PLMs is constructing a preference distribution $\mathcal{D}$ from scalar labels, as exhaustively pairing labeled sequences does not scale.

\subsection{Preference sampling for DPO with PLMs}

\paragraph{Output space partitioning.} A common strategy is to partition labeled sequences by thresholds defined by design objectives (e.g., "get sequences above $x$ stability") or quantiles, and assign sequences above as "preferred" and sequences below as "dispreferred" \cite{mistani2024preference, xue2025improving, xu2025protein, rong2025enerbridge, xue2025improvingproteinsequencedesign}. In terms of complexity, subsampling schemes, such as fixing $k$ losers per winner, can be used and scale effectively with dataset size. Although this aligns well for coarse optimization goals, treating all pairs within the same partition as equivalent reduces resolution among high-performing sequences, which is limiting in late-stage optimization. 

\paragraph{Rank-space sampling.}
Unlike thresholding, which collapses labels into binary partitions, rankings preserve relative relationships across all sequences. Methods such as RRHF \cite{yuan2023rrhf} and PRO \cite{song2024preference} show that training on rankings (or their induced pairs) leads to a more stable convergence in aligning LMs with human preferences. Applied to proteins, CtrlProt \cite{liu2025controllable} shows that a rank-wise preference objective improves controllability over pairwise methods for multi-objective optimization. Widatalla et al. \cite{widatallaAligningProteinGenerative2024} evaluate a `gap level' heuristic that down-samples pairs of sequences at fixed distances in rank space, and find that most gap levels have comparable performance to randomly sampling pairs. This suggests that not all pairs are equally informative and that randomly pruning pairs in the output space risks diluting the training signal.

\paragraph{Informative pairs.}

Other methods have been proposed to prioritize preference pairs that provide a stronger training signal. Maru et al. \cite{maru2025learning} propose a hybrid sampling strategy that mixes global and local perturbations, selecting pairs with large predicted stability differences for the former and pairs ten mutations apart for the latter. They find no improvement over randomly selecting pairs. Beyond proteins, other strategies have been proposed to improve informativeness, such as stratified sampling to ensure coverage across score bins, oversampling rare but informative regions, or curriculum learning to gradually introduce harder comparisons as the model improves \cite{lin2025activedpo, pattnaik2024curry, croitoru2025curriculum, cai2025k, hou2025novo}. 

\section{Methods: Group-DPO (g-DPO)}

In NLP, preferences are inherently structured: multiple outputs are generated from the same prompt, and annotators decide which is better \cite{ouyang2022training, bai2022training}. This provides a shared context that makes likelihood ratios between outputs meaningful. On the other hand, comparing outputs from different prompts yields likelihoods ratios that are dominated by prompt differences rather than output preference. Proteins have a related dynamic: even when variants come from the same wild type, comparing distant variants conflates the effects of many mutations, collapsing into coarse preference signals, whereas local comparisons better capture subtle, non-additive effects \cite{johnson2023epistasis, hayashi2006experimental}. Our work, therefore, leverages proximity in sequence space to construct preference pairs, enabling the model to learn fine-grained mutational effects while reducing the number of training pairs.

Our g-DPO training framework is conducted in three stages. We begin with unsupervised fine-tuning of a PLM on evolutionarily related sequences of a wild type protein (evo-tuning) \cite{alley2019unified, gordon2024protein}. Next, given an experimental mutant dataset of the relevant wild type, we apply union mask clustering, a greedy agglomerative procedure that groups sequences based on shared mutational positions. 
\begin{wrapfigure}{r}{0.32\textwidth}  
  \centering
  \vspace{-5pt}
  \includegraphics[width=\linewidth]{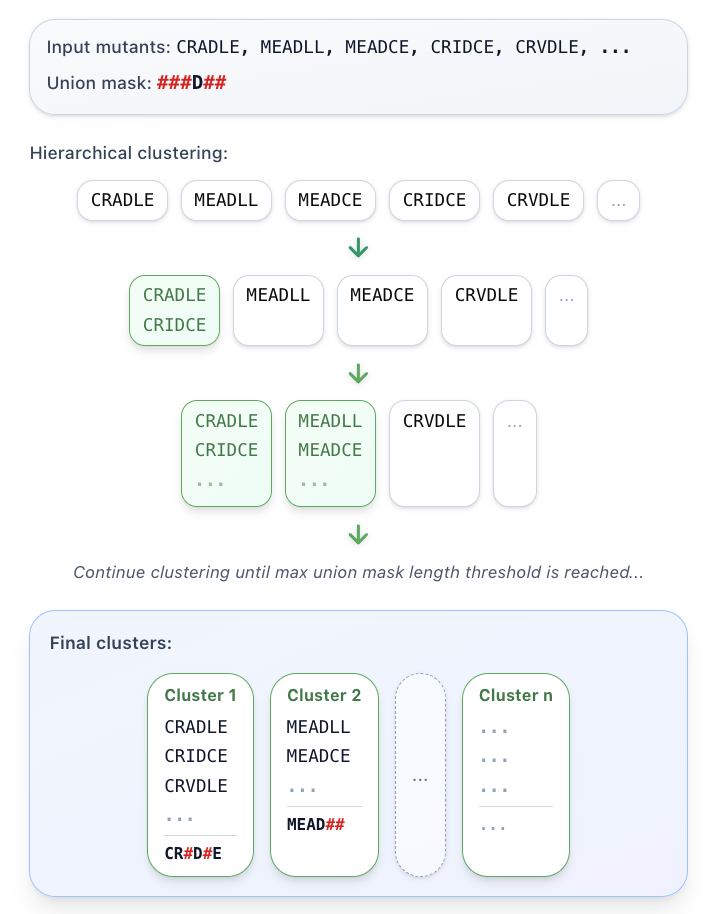}
    \caption{Clustering of mutants with overlapping mutations.}
  \label{fig:clustering}
  \vspace{-40pt}
\end{wrapfigure}
Finally, we sample groups of sequences from each cluster and evaluate the DPO loss over all preferences within each group to further fine-tune our evo-tuned model. Although some assay variants could appear within the evo-tuning training data, this does not constitute data leakage as evo-tuning uses no assay labels. This unsupervised design choice is intended to provide general evolutionary context to the model, opposed to labeled samples for model alignment.

\subsection{Union mask clustering}

Given a set of aligned sequences 
$S = (s^1, \ldots, s^n), \; s^i \in \mathcal{A}^L$, 
over the amino acid alphabet $\mathcal{A}$, define the \emph{union mask} as $M(S) = \{ i \in [\![ L]\!] : \exists j,k \ \text{such that } s_i^{(j)} \neq s_i^{(k)} \}$, and let $m(S) := |M(S)|$ denote the size of the union mask. We cluster sequences with a greedy agglomerative procedure with a linkage that prefers merges that keep the union masks small:

\begin{enumerate}\setlength{\itemsep}{0pt}\setlength{\itemindent}{-1.5em}
  \item \textbf{Initialization.}  
  Let $\mathcal{C} \leftarrow \{ \{s^1\}, \ldots, \{s^n\} \}$.  
  For each cluster $C \in \mathcal{C}$, store its union mask $M(C)$. 

  \item \textbf{Linkage.}  
  For clusters $C_i, C_j$, define the cost of merging $C_j$ into $C_i$ as
  \[
  \phi(C_i, C_j) = m(C_i \cup C_j) - m(C_i) 
  = \big| M(C_i) \cup M(C_j) \cup M(\{s, s'\}) \big| - m(C_i),
  \]
  where $s \in C_i$ and $s' \in C_j$ are arbitrarily chosen.

  \item \textbf{Greedy merge.}  
  Repeatedly select 
  \[
  (C_p, C_q) = \arg\min_{i \neq j} \phi(C_i, C_j),
  \]
  breaking ties by minimum $m(C_i)$.  
  Replace clusters $C_p$ and $C_q$ by $C_p \cup C_q$, and store the new union mask $M(C_p \cup C_q) = M(C_p) \cup M(C_q) \cup M(\{s, s'\})$. 

  \item \textbf{Stopping.}  
  Stop when the next best merge would violate the union mask ratio threshold $\tau \in [0,1]$,
  \[
  \min_{i \neq j} \phi(C_i, C_j) > \tau L.
  \]
\end{enumerate}

\paragraph{Time complexity.} Calculating $\phi(C_i, C_j)$ is $O(L)$, since $M(C_i)$ and $M(C_j)$ are stored, and only $M({s, s'})$ must be computed. In each merging step, computing all costs $\phi$ is $O(n^2L)$. Using a heap over cluster pairs, the final time complexity is $O(n^2 \log n + n^2 L)$. When $n$ is large, we first coarse-cluster with MMseqs2 \cite{steinegger2017mmseqs2}, which provides approximate linear-time clustering for large datasets, and then run union mask clustering independently within those buckets.

\subsection{Group sampling} 

\paragraph{Log likelihood approximation.} Evaluating the DPO loss on a pair $(y_w, y_l)$ requires likelihoods for both sequences. For masked language models (MLMs) like ESM-2 \cite{lin2023evolutionary}, computing pseudo log-likelihood (PLL) \cite{salazar2019masked} scores has become the standard way to approximate sequence likelihoods. This requires one forward pass per position and is therefore expensive. To reduce this cost, we exploit the union mask $D = M(\{y_w, y_l\})$, and create a jointly masked input $y_{\setminus D}$, where all differing positions between $y_w$ and $y_l$ are masked. One forward pass on $y_{\setminus D}$ provides logits for all positions. We then approximate $\log p(y_w) \approx \sum_{i \in D} \log p({y_w}_i|y_{\setminus D})$ (same for $y_l)$. Positions outside $D$ do not need to be evaluated since their logits agree. This is equivalent to a mean-field approximation, where we assume that, conditional on the observed tokens, the masked positions are independent. Although approximate, it is reliable when the differing positions are few compared to the sequence length, since the error scales linearly with the size of the difference mask. Similar multi-mask approximations have already been applied successfully in prior work \cite{meier2021language, chang2022maskgit, nisonoff2024unlocking, zheng2024masked}.

Since union-mask clustering controls the union-mask size of clusters, we can extend this approximation beyond pairs. We uniformly sample without replacement groups of $g$ sequences from each cluster. We can then approximate the likelihood of each sequence in a group with a single forward pass. We then evaluate the DPO loss over all pairwise comparisons inside the group. Sampling continues until every sequence has appeared in at least one group, which defines one training epoch.

\section{Experiments}

\begin{figure}[t]
    \centering
    \vspace{-5pt}
    \begin{subfigure}[t]{0.32\textwidth}
        \includegraphics[width=\linewidth]{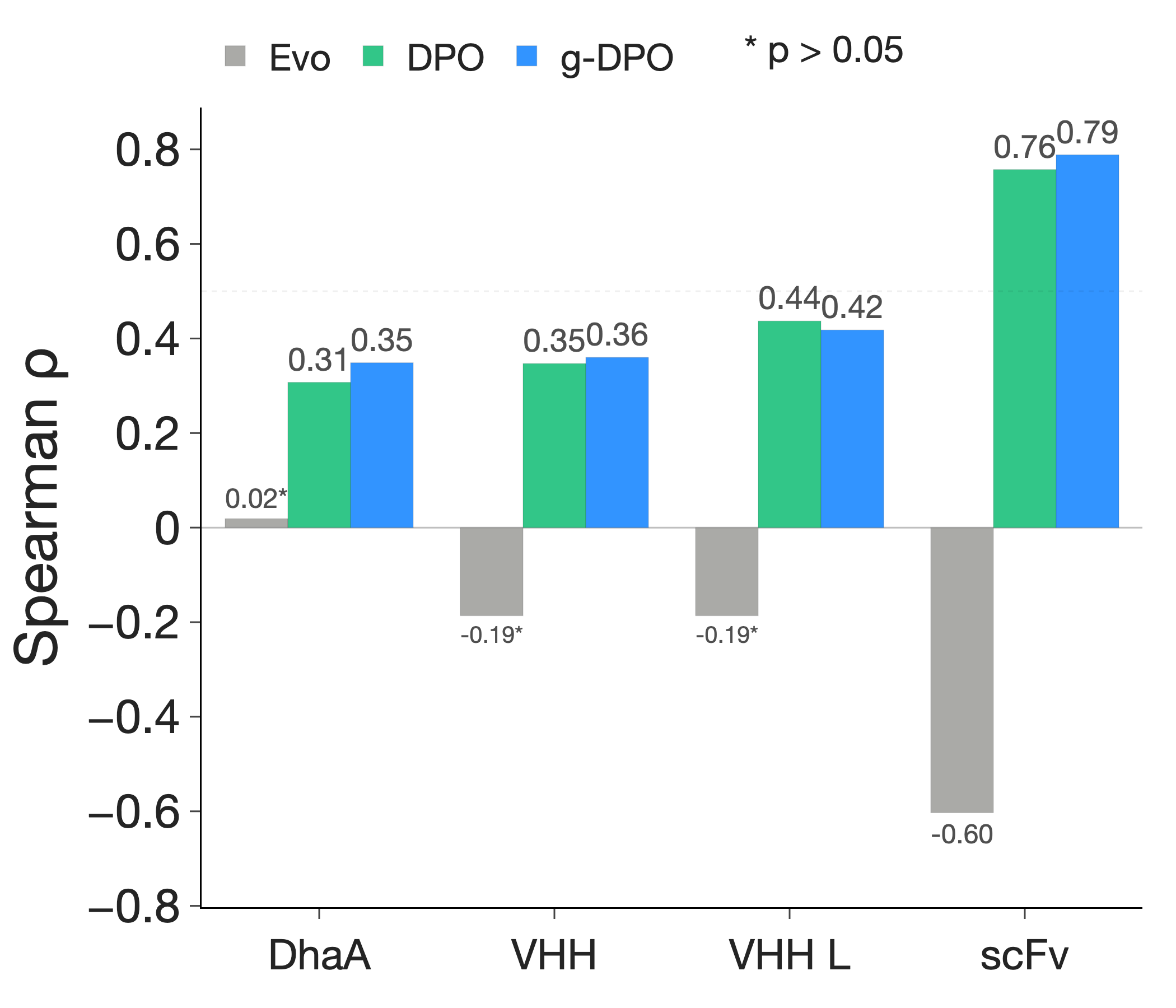}
        \subcaption*{(a)}
    \end{subfigure}\hfill
    \begin{subfigure}[t]{0.22\textwidth}
        \includegraphics[width=\linewidth]{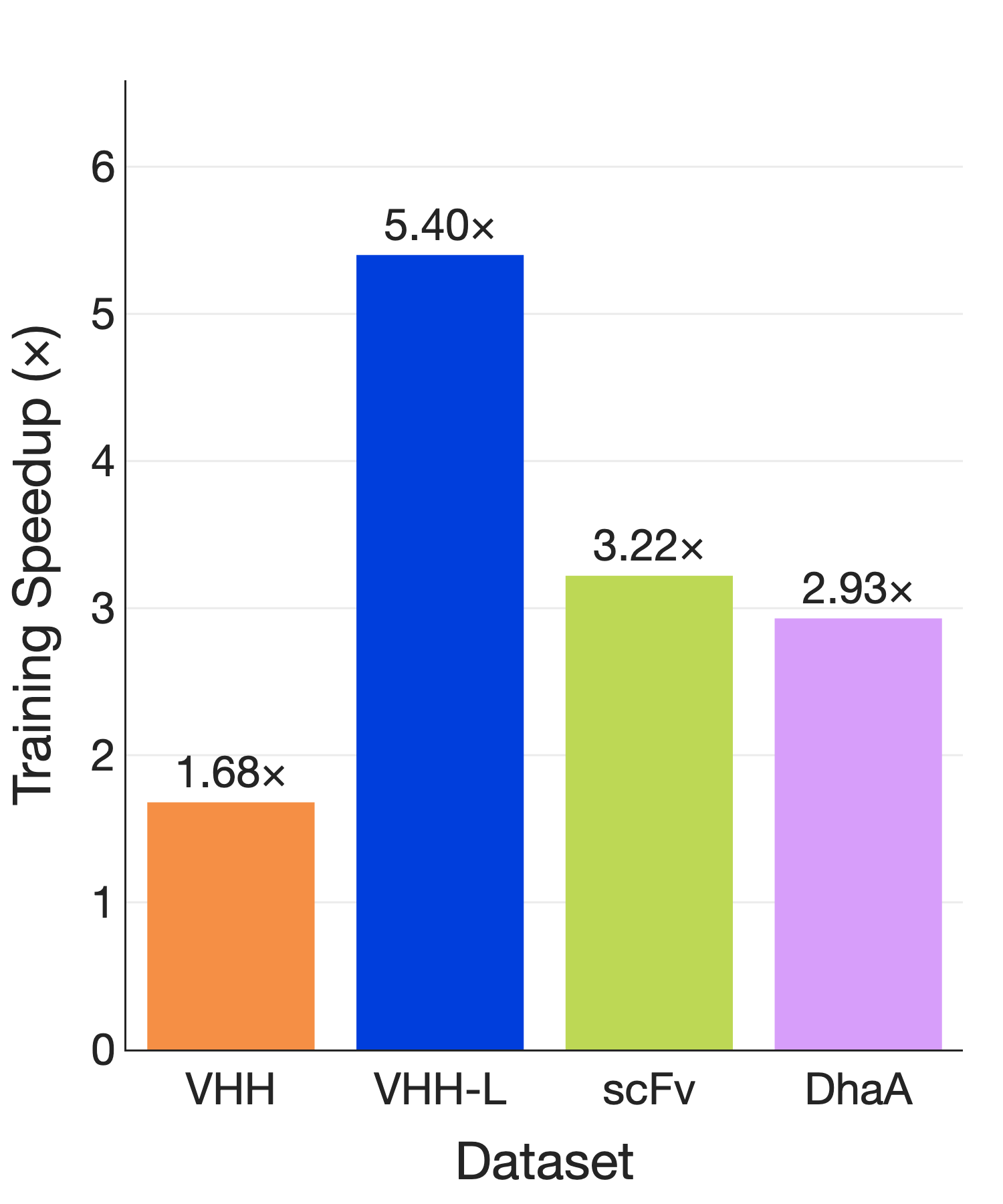}
        \subcaption*{(b)}
    \end{subfigure}\hfill
    \begin{subfigure}[t]{0.4\textwidth}
        \begin{subfigure}[t]{0.32\linewidth}
            \includegraphics[width=\linewidth]{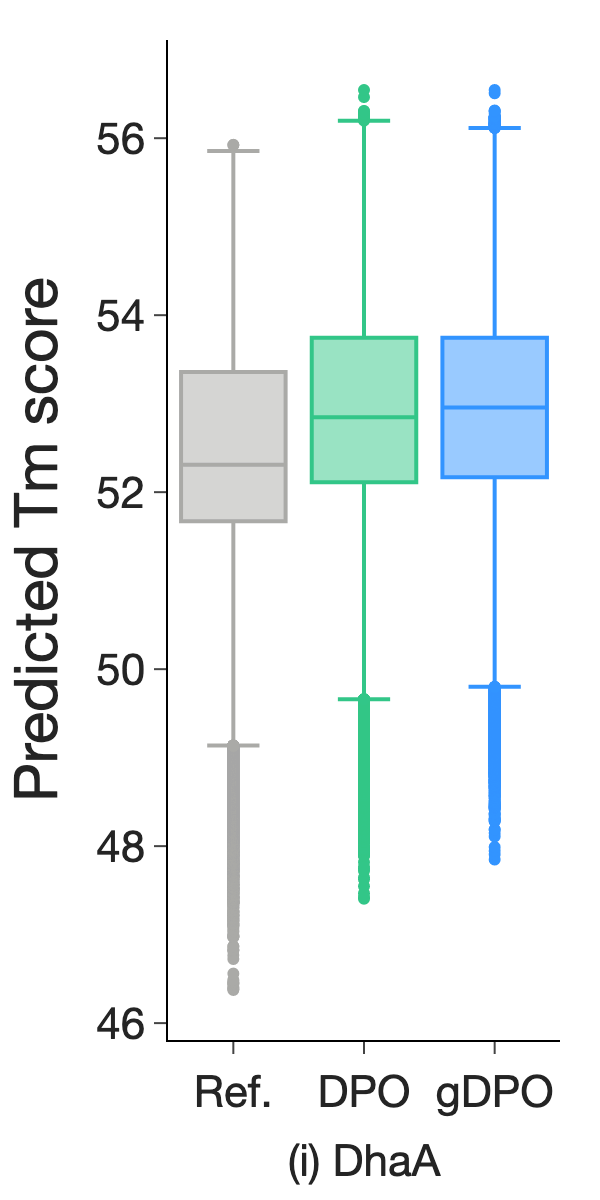}
        \end{subfigure}
        \begin{subfigure}[t]{0.32\linewidth}
            \includegraphics[width=\linewidth]{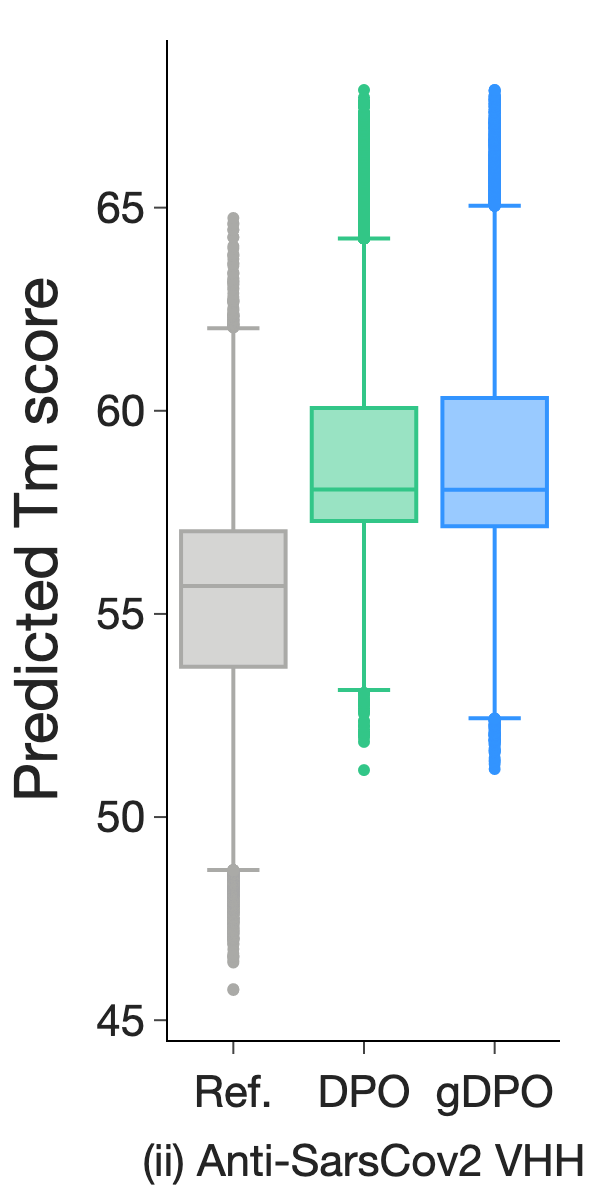}
        \end{subfigure}
        \begin{subfigure}[t]{0.32\linewidth}
            \includegraphics[width=\linewidth]{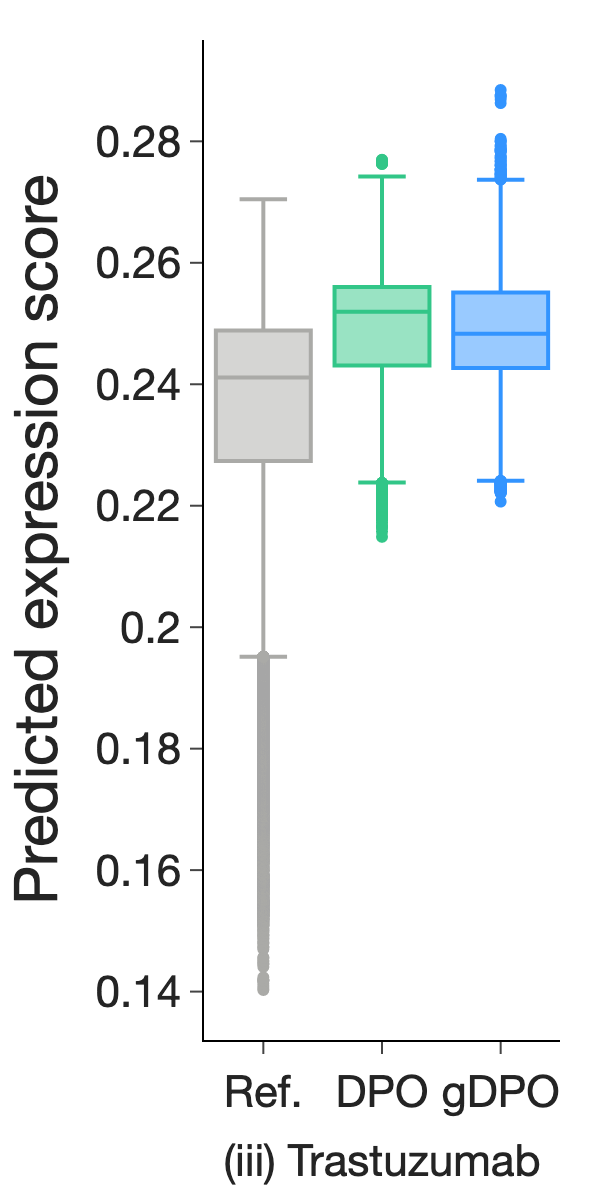}
        \end{subfigure}
        \subcaption*{(c)}
    \end{subfigure}
    \caption{\textit{In silico} evaluation of g-DPO. (a) Spearman correlation ($\rho$) of evo-tuned, DPO, and g-DPO models. (b) Convergence speedup of g-DPO vs. DPO (higher better), measured on a single NVIDIA A100 GPU; wall-clock times are in \textbf{Table~\ref{tab:runtime_comparison}}. (c) Predicted property distributions of sequences generated from the evo-tuned, DPO, and g-DPO models using beam search. KS tests confirm that g-DPO and DPO yield nearly identical, and improved distributions over the reference.}
    \label{fig:insilico}
    \vspace{-10pt}
\end{figure}

Evaluating generative models for protein optimization is challenging and ultimately requires \textit{in vitro} validation. We test whether g-DPO (i) preserves or improves model quality relative to DPO, (ii) reduces training cost, and (iii) whether \emph{in silico} gains translate to experimental outcomes. We evaluate g-DPO on experimentally measured variants from three wild-type proteins:
\begin{table}[h]
\centering
\scriptsize                          
\renewcommand{\arraystretch}{0.88}   
\setlength{\tabcolsep}{5pt}          
\begin{tabular}{lccc}
\toprule
\textbf{Dataset} & \textbf{Function} & \textbf{$N$} & \textbf{Positional coverage} \\
\midrule
Anti-SARS-CoV-2 VHH               & Thermostability & 462  & 47.1\% \\
Anti-SARS-CoV-2 VHH L             & Thermostability & 1833 & 92.4\% \\
Trastuzumab scFv & Expression      & 76   & 13.1\% \\
Haloalkane dehalogenase (DhaA)    & Thermostability & 474  & 40.3\% \\
\bottomrule
\end{tabular}
\end{table}

For VHH L we report training metrics and one \textit{in silico} metric, as it is included for scaling analysis. We use ESM-2-650M \cite{lin2023evolutionary} as our pre-trained model; further details are provided in Appendix~\ref{expdetails}.

\subsection{\textit{In silico} validation}

For each dataset, we report 
(i) rank correlation between PLL scores and experimental measurements on holdout test sets, which tests whether model likelihoods align with ground truth preferences;
(ii) convergence time to early stopping, to measure training efficiency; and 
(iii) generative quality, measured by scoring sequences generated with beam search \cite{sutskever2014sequence}. \begin{wrapfigure}{r}{0.55\textwidth}  
  \centering
  \begin{minipage}{0.9\linewidth}
  \begin{tabular}{ c @{\quad} c }
    \includegraphics[width=0.49\linewidth]{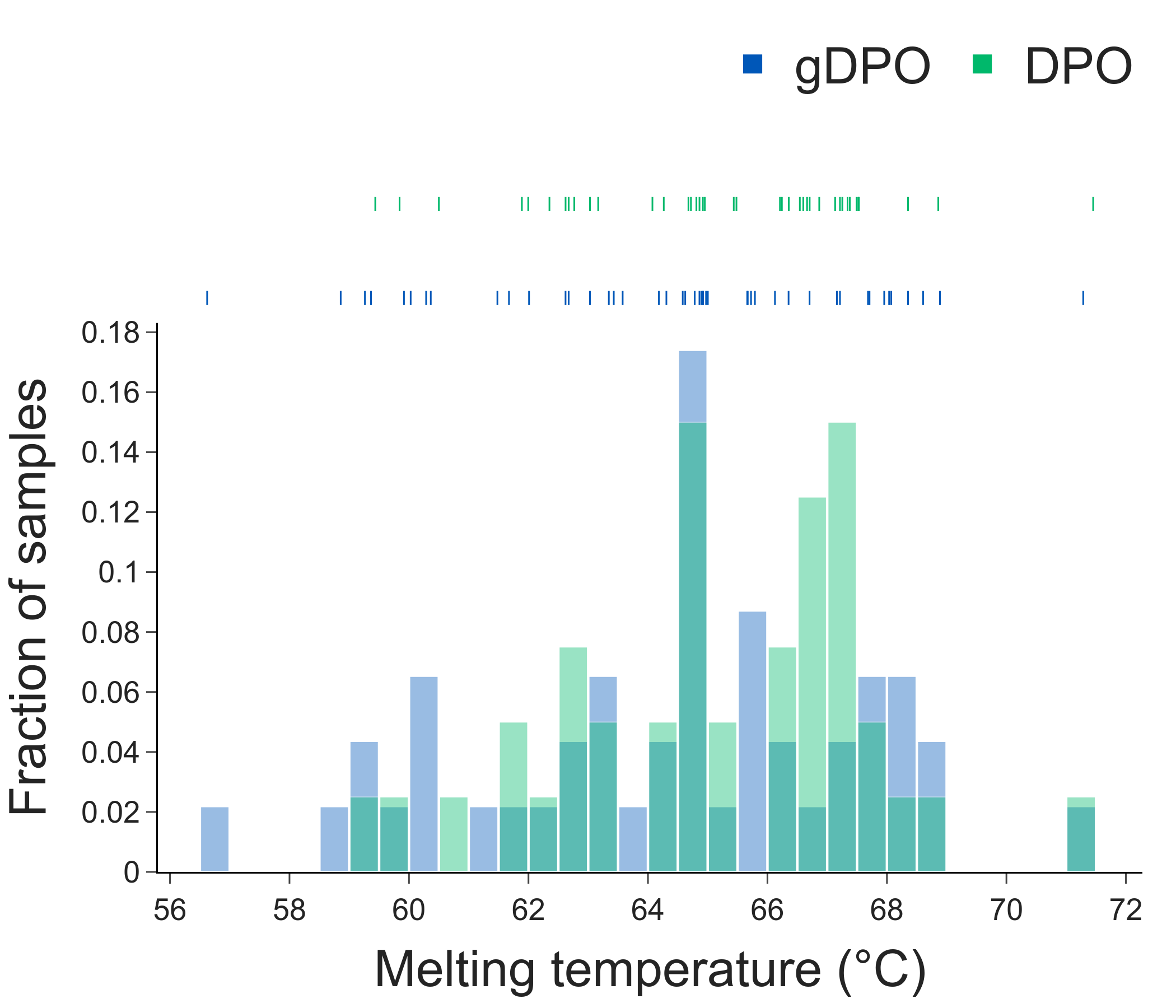} &
    \includegraphics[width=0.49\linewidth]{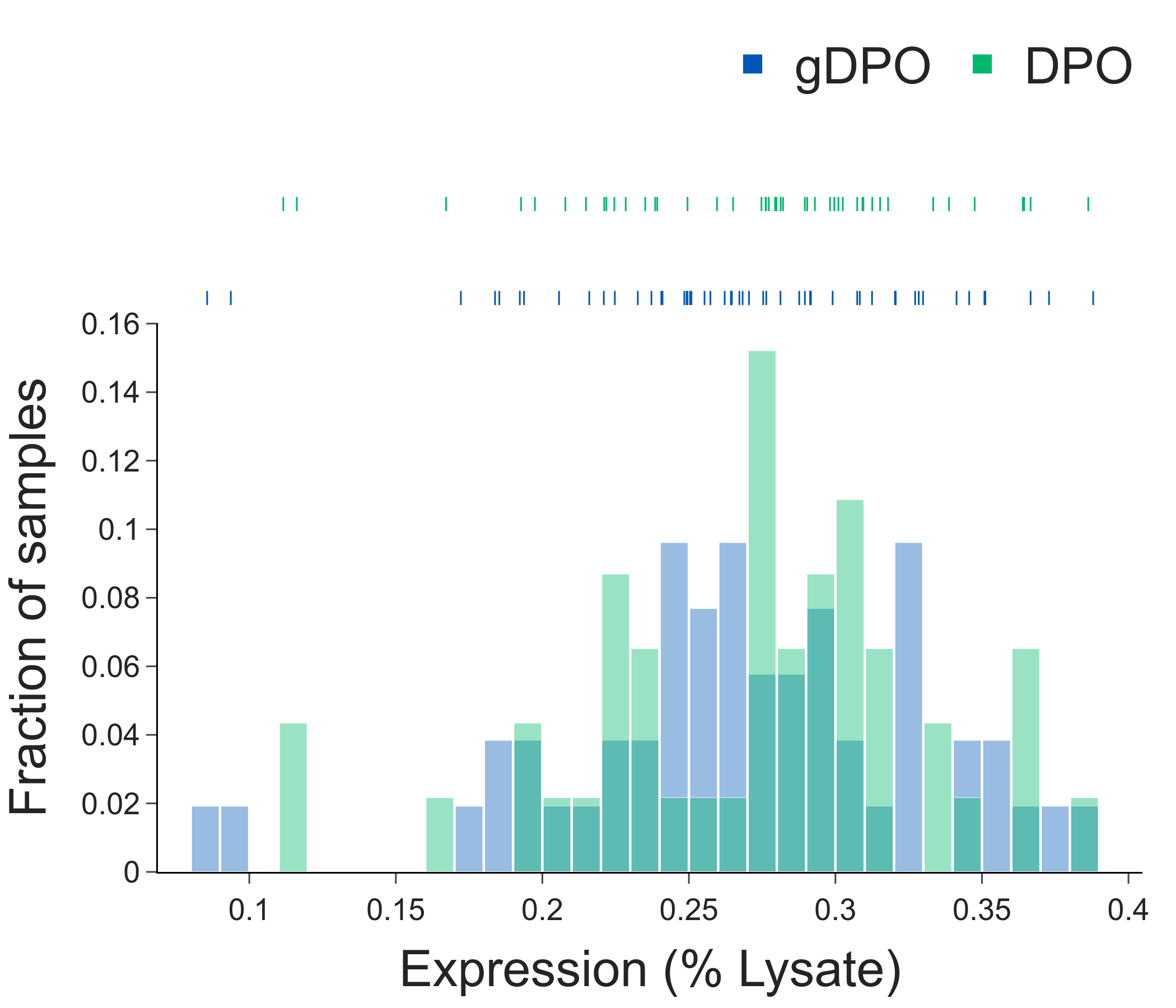} \\
    \small (a) & \small (b)
  \end{tabular}
  \caption{\textit{In vitro} validation. Distribution of assay measurements for sequences designed with g-DPO and DPO on (a) thermostability of DhaA and (b) expression of Trastuzumab. Distributions show that both models yield comparable outcomes.}
  \label{fig:invitro}
  \end{minipage}
  \vspace{-15pt}
\end{wrapfigure}We score the generated sequences with independent predictors trained on the same experimental datasets.

Both DPO and g-DPO improve rank correlation relative to the evo-tuned reference model (Figure~\ref{fig:insilico}a) and shift the distribution of generated sequences toward better predicted function (Figure~\ref{fig:insilico}c).
 The distributions produced by DPO and g-DPO are statistically indistinguishable, indicating that g-DPO preserves generative quality while reducing training cost (Figure~\ref{fig:insilico}b). Statistical significance was assessed with two-sample KS tests, with full results reported in Table \ref{ks-table}.

\subsection{\textit{In vitro} validation} 

To test whether \textit{in silico} improvements translate to experimental outcomes, we validated a subset of generated sequences in the lab for two optimization tasks: thermostability of DhaA and expression of Trastuzumab. Candidates were down-selected from the \textit{in silico} pool using a criterion that averages performance across the top-$k$ predicted sequences ($k=3$), following the Monte Carlo strategy of DiscoBAX \cite{lyle2023discobax}. Assay results confirm that DPO and g-DPO yield comparable outcomes, with no significant differences between their distributions.

\subsection{Ablations}
We ablate clustering ($\tau$) and grouping ($g$) to understand their individual effects and how they interact. All ablations are reported on the Anti-SARS-CoV-2 VHH dataset. 

\begin{wrapfigure}{r}{0.56\textwidth}  
  \vspace{-10pt}  
  \centering
  \begin{subfigure}[t]{0.55\linewidth}
      \includegraphics[width=\linewidth]{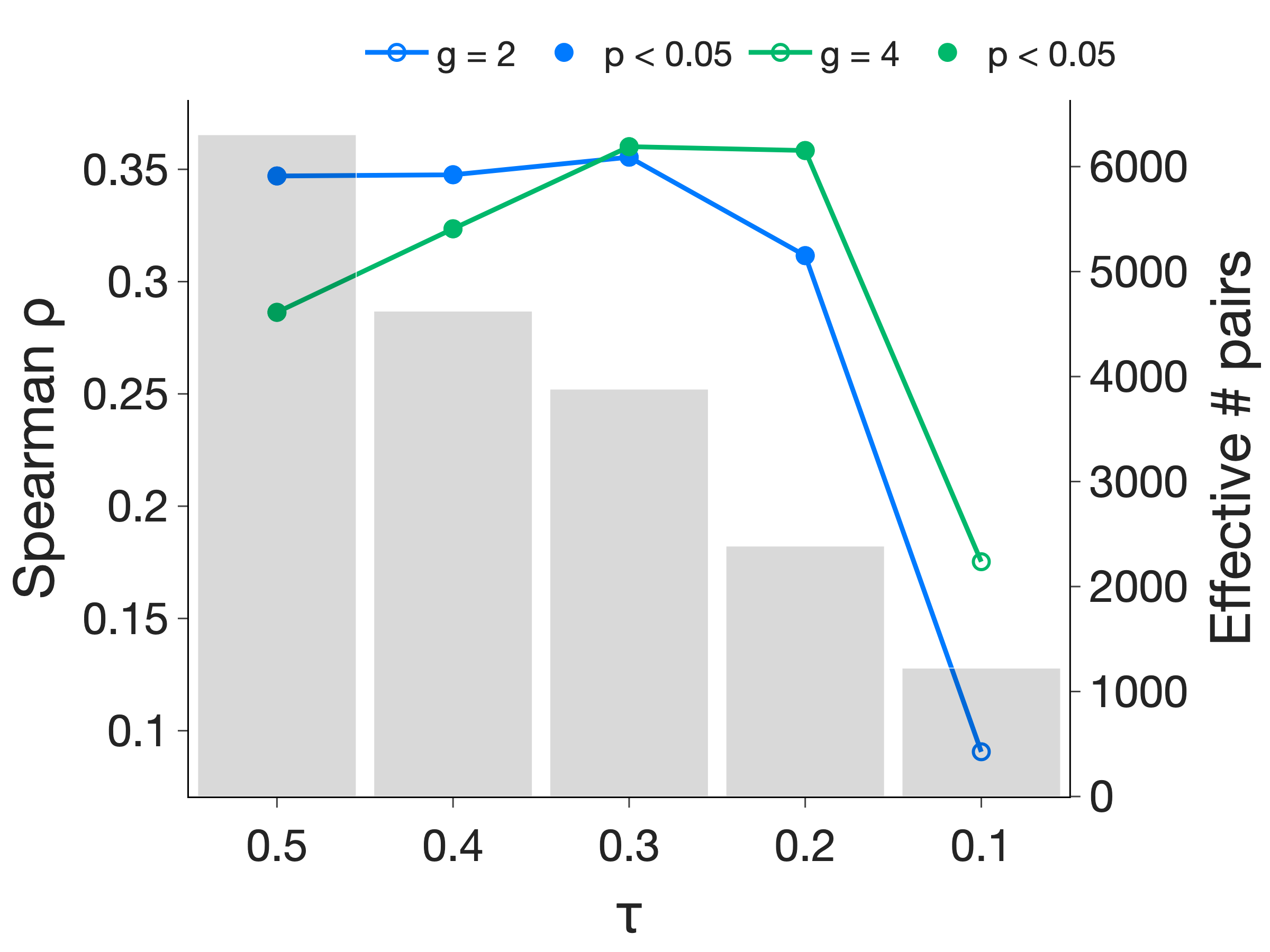}
      \caption{}  
  \end{subfigure}\hfill
  \begin{subfigure}[t]{0.42\linewidth}
      \includegraphics[width=\linewidth]{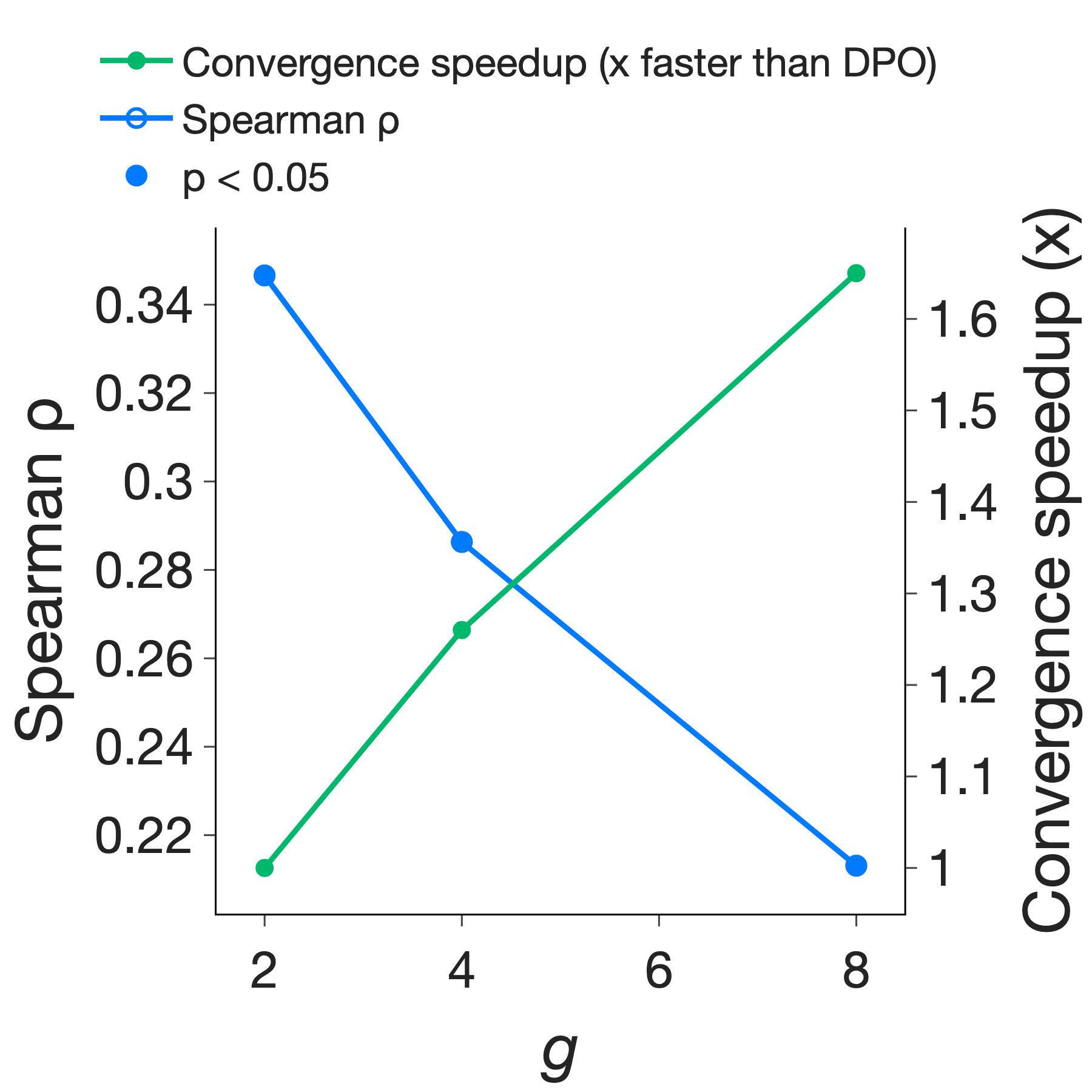}
      \caption{}  
  \end{subfigure}
  \caption{(a) Effect of clustering threshold $\tau$ on performance and training pairs. 
           (b) Effect of grouping size $g$ on performance and convergence speed without clustering.}
  \label{fig:spearman_all}
  \vspace{-10pt}
\end{wrapfigure}

\paragraph{Clustering, no grouping.} We sweep the clustering threshold $\tau$ from $0.5 \rightarrow 0.1$, where a smaller $\tau$ yields more clusters, reduces the number of training pairs and speeds up convergence. In Figure~\ref{fig:spearman_all}a with $g=2$, we see that as $\tau$ decreases to $\approx 0.3$, model performance remains unchanged. Beyond this threshold, performance declines as clusters become too tight and useful signal is lost. This indicates that many pairs are redundant and can be safely pruned in input space up to a moderate threshold ($\tau \approx 0.3$).

\paragraph{Grouping, no clustering.} Grouping amortizes likelihood computations by sharing a union mask and a single forward pass across $g$ sequences, but comes at a cost. When mutations span a large fraction of the sequence, union masks become too large, likelihoods are overly approximated, and model performance drops (Figure~\ref{fig:spearman_all}b). Hence, grouping alone is insufficient when the mutation span is high. 

\paragraph{Clustering with grouping.} Combining clustering with grouping ($g=4$ in Figure~\ref{fig:spearman_all}a) yields the expected trade-off. With little to no clustering, grouping degrades model performance, as previously mentioned. As the clustering threshold increases, performance improves and eventually matches that of $g=2$ at $\tau \approx 0.3$, while convergence is significantly faster for $g=4$. Beyond $\tau \approx 0.3$, performance drops as clustering becomes too strict and valuable training signal is lost. This drop is less severe for $g=4$, probably because within-group preferences are processed in one batch, with coupled likelihoods and update steps, while with $g=2$, they can be split across batches. Practically, clustering controls the training signal (removing redundant pairs), and grouping controls efficiency; however, accurate grouping is dependent on a reasonable clustering level for the likelihood approximation to hold.

\section{Conclusion, limitations and future work}
We present g-DPO, a scalable variant of DPO for PLMs, that combines sequence-space clustering, to prune redundant preference pairs, with grouped likelihood amortization, to reduce computation. Across three protein engineering tasks, g-DPO matches DPO performance on both \textit{in silico} metrics and \textit{in vitro} results, while converging up to 5.4× faster. In practice, moderate clustering with grouping yields the best trade-off; excessive clustering removes training signal, and grouping alone can degrade performance when mutation span is large and likelihood approximations break down. 

Our experiments span three mutational landscapes with 76-474 variants, representative of typical assay-driven protein engineering datasets. However, extending evaluation to larger and more diverse benchmarks (multiple wild types, broader mutation spans) would allow to further test the scalability and robustness of g-DPO. We expect the scalability advantage of g-DPO to become even more important in such settings. A natural extension is to move beyond a pairwise objective and explore rank-based objectives which could leverage the full ranking information of variants within clusters. Moreover, while this study focused on unimodal protein sequence PLMs, the clustering and grouping strategies are modality-agnostic. Extending g-DPO to multi-modal foundation models (e.g., sequence-structure or sequence-function) is a promising direction for future work.

\bibliographystyle{plainnat}

\bibliography{neurips_2025}

\appendix
\section{Preliminaries}

\subsection{Direct Preference Optimization (DPO) Overview}\label{dpoloss}

Given a pre-trained reference model $\pi_{\text{ref}}$, the goal of DPO is to learn a policy $\pi_\theta$ that maximizes the probability of generating preferred sequences over dispreferred ones. 

Consider a dataset of $N$ pairwise rankings, $D = \{ (x^{(i)}, y_w^{(i)}, y_l^{(i)}) \}_{i=1}^N$, where $x^{(i)}$ is the masked input and $y_w^{(i)}$ and $y_l^{(i)}$ are the preferred and dispreferred outputs. The preference distribution $p^*$ can be expressed given a Bradley-Terry reward model $r^*$ and a parameterization of the score $s(x, y) = \exp(r^*(x, y))$,

\[
p^*(y_w \succ y_l \mid x) = \frac{\exp(r^*(x, y_w))}{\exp(r^*(x, y_w)) + \exp(r^*(x, y_l))} 
= \sigma( r^* (x, y_w) - r^*(x, y_l))
\]

Under the RLHF framework, a parameterized reward model $r_\Phi (x, y)$ can be estimated to match the optimal reward function $\pi_*$, using the negative log-likelihood loss,

\[
L_R(r_\Phi, D) = - \mathbb{E}_{(x, y_w, y_l) \sim D} \,[ \log \sigma( r_\Phi (x, y_w) - r_\Phi (x, y_l)) ]
\]

This reward model can then be used to optimize the optimal policy, 

\[
\pi^* = \arg \max \mathbb{E}_{x \in D, y \sim \pi_\theta} [r_\Phi(x,y)]- \beta D_{\text{KL}} [\pi_\theta(y\mid x) \| \pi_{\text{ref}} (y\mid x)]
\]

The KL term is there to ensure that the outputs from the learned policy do not diverge too much from the original policy, as we assume that the fine-tuned base model already produces reliable outputs. Using Gibbs' inequality, we can derive the optimal solution to this optimization problem, 

\[
\pi^*(y\mid x) = \frac{1}{Z(x)} \pi_{\text{ref}} (y\mid x)e^{\tfrac{1}{\beta} r_\Phi (x,y)}
\]

where $Z(x) = \sum_y \pi_{\text{ref}} (y\mid x)e^{\tfrac{1}{\beta} r_\Phi (x,y)}$, an untractable term. We can reorganize this term to solve for $r_\Phi$ or $r^*$ with its corresponding policy $\pi^*$, which we can plug back into (2) to obtain,

\[
\mathcal{L}_{\text{DPO}}(\pi_\theta; \pi_{\text{ref}}) =
- \mathbb{E}_{(y_w, y_l) \sim D} \left[
\log \sigma \big( \beta \log \tfrac{\pi_\theta(y_w)}{\pi_{\text{ref}}(y_w)} 
- \beta \log \tfrac{\pi_\theta(y_l)}{\pi_{\text{ref}}(y_l)} \big)\right]
\]

where $\sigma$ is the sigmoid function and $\beta$ is a scaling parameter. 

This objective encourages the policy to increase the likelihood of preferred sequences relative to dispreferred ones, while the reference model serves as a regularizer to prevent the policy from deviating too far from the original distribution.

\section{g-DPO Details}

\subsection{Union mask clustering algorithm}

\begin{algorithm}[H]
\caption{Greedy Union-Mask Clustering}
\label{alg:union-mask-linkage}
\begin{algorithmic}[1]
\Require Sequences $\{s^1,\dots,s^n\}, s^i \in \mathcal{A}^L$; length $L$; union-mask ratio threshold $\tau \in [0,1]$.
\Ensure A set of clusters $\mathcal{C}$.

\State \textbf{Initialization:} $\mathcal{C} \gets \big\{\{s^1\}, \dots, \{s^n\}\big\}$.
\ForAll{$C \in \mathcal{C}$}
  \State Store its union mask $M(C)$ and its size $m(C) = |M(C)|$.
\EndFor

\Function{$\phi$}{$C_i, C_j$} \Comment{Merge cost of $C_j$ into $C_i$}
  \State Choose arbitrary $s \in C_i$, $s' \in C_j$.
  \State \Return $|\,M(C_i) \cup M(C_j) \cup M(\{s,s'\})\,| - m(C_i)$.
\EndFunction

\While{true}
  \State $(p,q) \gets \arg\min_{i \neq j} \big(\phi(C_i, C_j)\big)$ \Comment{Tie-break: smaller $m(C_i)$}
  \State $c^\star \gets \phi(C_p, C_q)$
  \If{$c^\star > \tau L$} \Comment{Stopping rule}
    \State \textbf{break}
  \EndIf
  \State Choose arbitrary $s \in C_p$, $s' \in C_q$.
  \State $C_{\text{new}} \gets C_p \cup C_q$
  \State $M(C_{\text{new}}) \gets M(C_p) \cup M(C_q) \cup M(\{s,s'\})$
  \State $m(C_{\text{new}}) \gets |M(C_{\text{new}})|$
  \State $\mathcal{C} \gets \big(\mathcal{C} \setminus \{C_p, C_q\}\big) \cup \{C_{\text{new}}\}$
\EndWhile

\State \Return $\mathcal{C}$
\end{algorithmic}
\end{algorithm}

\subsection{Single forward pass log-likelihood approximation algorithm}

The DPO loss requires log-likelihood differences between sequences. By constructing a union mask for a group $G$, all sequences within $G$ are evaluated under the same masked context, making their likelihoods directly comparable. This ensures that the log-likelihood ratios used in the loss are valid. Importantly, the approximation only holds for sequences scored under the same union mask.

 \begin{figure}[h]
\centering
\includegraphics[width=0.8\textwidth]{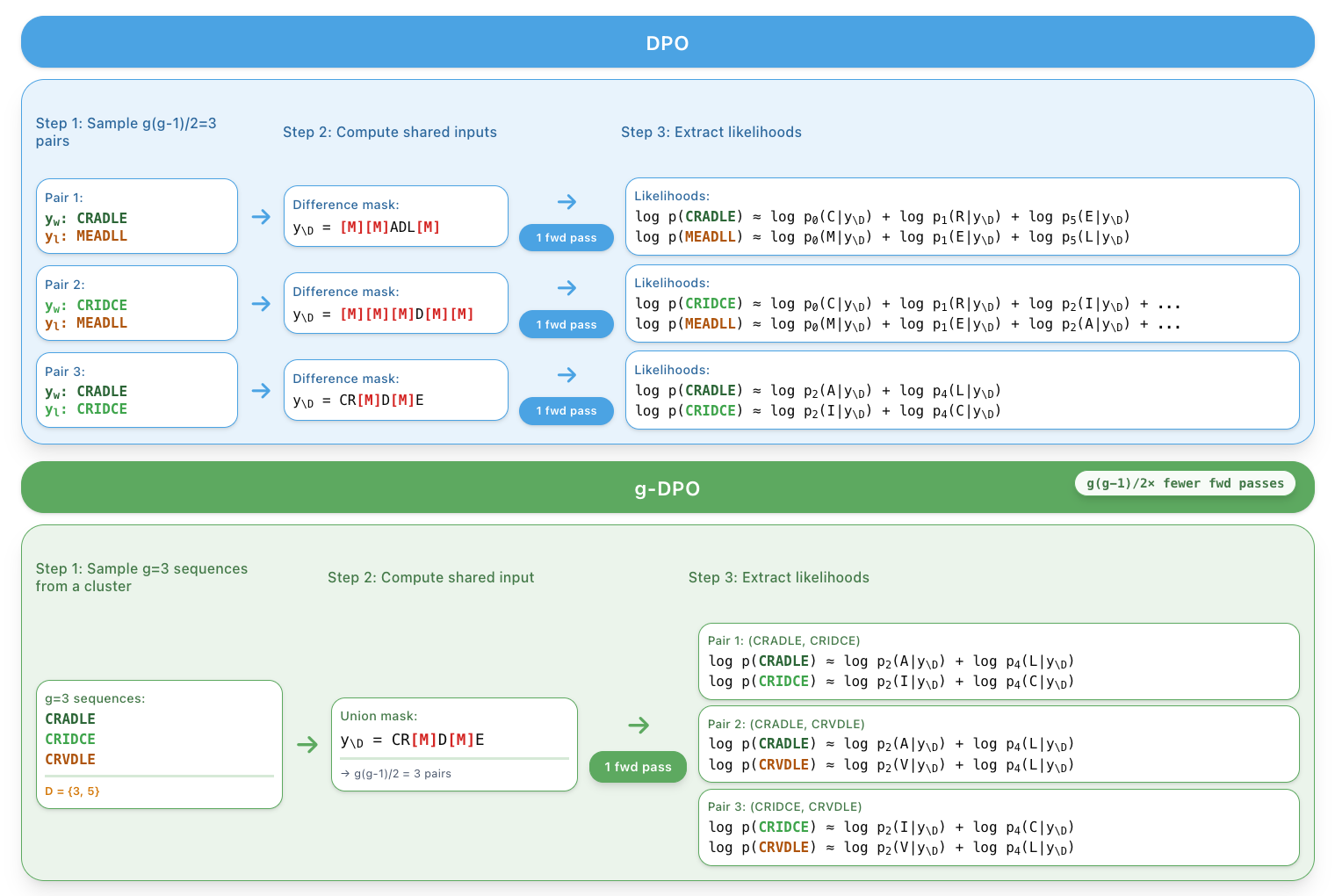}
\caption{Likelihood computation comparison between g-DPO and DPO. }
\label{kendall_main}
\end{figure}

\begin{algorithm}[H]
\caption{Union-mask LL approximation}
\label{alg:group_pll}
\begin{algorithmic}[1]
\Require Cluster $C \subset \mathcal{C}$; group $G \subset C$ of size $g$; MLM model $\pi$; mask token \texttt{[MASK]}
\Ensure Per-sequence approximate likelihoods $\{\widetilde{\log p_\pi}(y)\}_{y \in G}$

\Function{UnionMask}{$G$}
  \State \Return $D \gets \{\, p \in [\![ L]\!] \mid \exists\, y,y' \in G:\; y_p \neq y'_p \,\}$
\EndFunction

\Function{Mask}{$y, D$}
  \State $x \gets y$; \quad $x_p \gets \texttt{[MASK]}$ for all $p \in D$
  \State \Return $x$
\EndFunction

\Function{Likelihoods}{$G, \pi$}
  \State $D \gets \Call{UnionMask}{G}$
  \State Choose arbitrary $y \in G$
  \State $x \gets \Call{Mask}{y, D}$
  \State $\mathrm{logits} \gets \pi(x)$ \Comment{single forward pass; returns token logits for all positions}
  \ForAll{$y \in G$}
    \State $\widetilde{\log p_\pi}(y) \gets \sum_{p \in D} \log \mathrm{softmax}(\mathrm{logits}_p)[\,y_p\,]$
  \EndFor
  \State \Return $D,\ \{\widetilde{\log p_\pi}(y)\}_{y \in G}$
\EndFunction

\end{algorithmic}
\end{algorithm}

\subsection{Training speedup}
g-DPO improves training efficiency in two ways. (1) Clustering prunes pairs that are redundant or less informative, reducing the quadratic growth of pairs and focusing updates on local neighborhoods that provide stronger training signal. (2) Grouping amortizes likelihood evaluations: for a group of size $g$, all $\binom{g}{2}$ pairwise preferences are computed from a single forward pass under the shared union mask, rather than one forward pass per pair. For example, with $g=4$, $6$ pairs are obtained from one forward pass instead of $6$ forward passes. These two mechanisms together reduce both the number of pairs processed and the cost per pair, resulting in $1.7$--$5.4\times$ faster convergence in our experiments. As datasets grow, the number of redundant pairs increases, but clustering should discard them. Consequently, the relative efficiency gain of g-DPO compounds with dataset size and with the structure of the underlying mutational landscape.

\section{Experiment details}\label{expdetails}

\subsection{Training details}\label{trainingdetails}

We start with pre-trained ESM-2-650M (weights are open-sourced) \cite{lin2023evolutionary}, which we fine-tune on evolutionarily related sequences retrieved via MMseqs2 \cite{steinegger2017mmseqs2} searches against ColabFold databases \cite{mirdita2022colabfold}. We follow the evo-tuning framework defined by \cite{alley2019unified}. Then, this model is further fine-tuned using the g-DPO framework with SGD using a learning rate of $7 \times 10^{-4}$ and no weight decay, with a 300-step linear warmup. The DPO loss is computed with $\beta = 0.04$ with a batch size of 64. Sequences were clustered with a maximum union mask size of $0.3L$, and groups of $g=4$ were sampled from each cluster. The loss was applied over all within-group pairs. This was the final configuration, though we ran hyperparameter sweeps and expect the optimal group size to depend on dataset scale and clustering. Validation loss was monitored every 250 steps, with checkpointing on validation loss and relative early stopping on validation loss that checks there has not been more than a 1\% improvement with a patience of 3 validation intervals. Training was performed on a single NVIDIA A100 GPU.

\subsection{Datasets}\label{datasetdetails}

\begin{figure}[H]
\centering
\begin{subfigure}[t]{0.3\textwidth}
  \includegraphics[width=\linewidth]{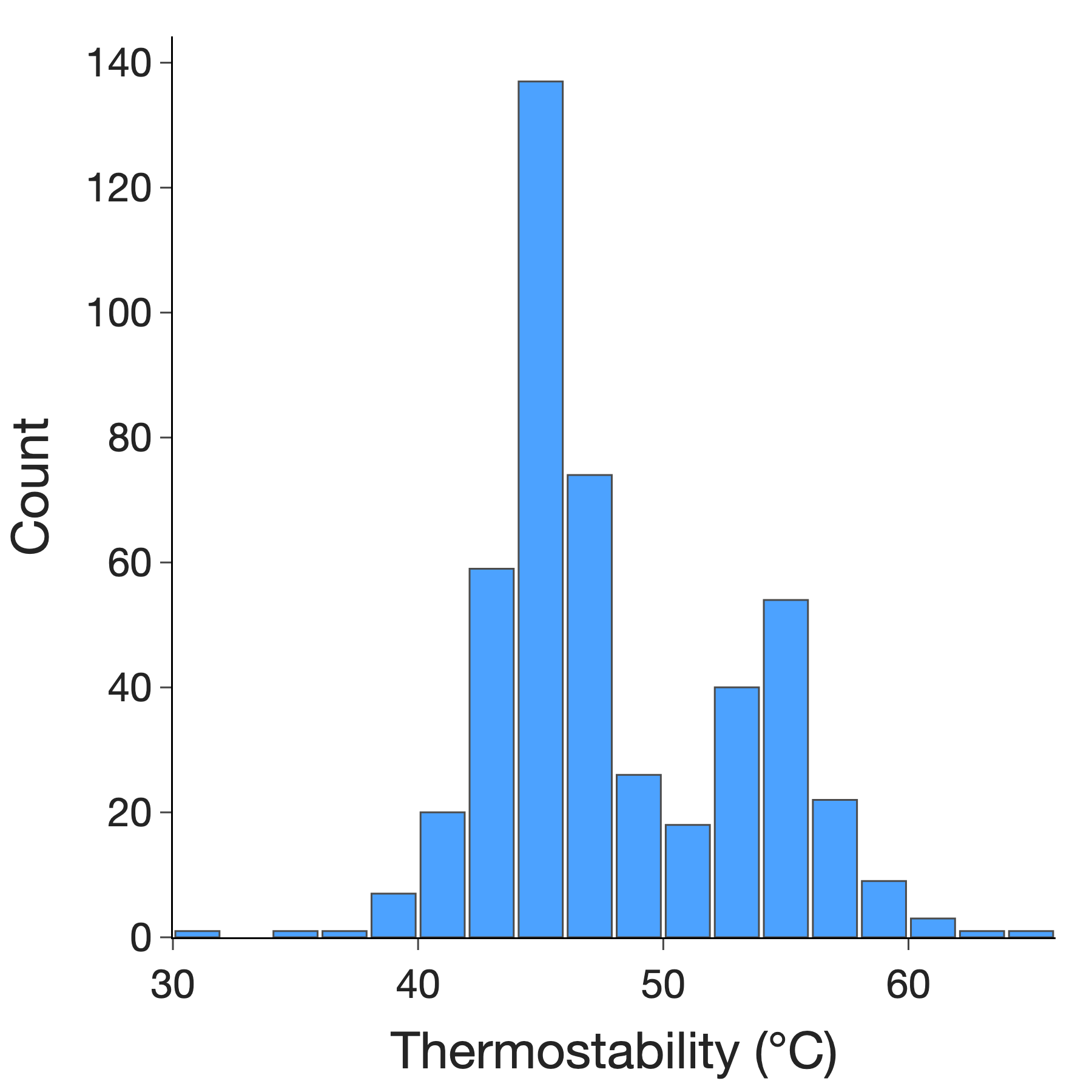}
  \caption{DhaA}
\end{subfigure}
\hfill
\begin{subfigure}[t]{0.3\textwidth}
  \includegraphics[width=\linewidth]{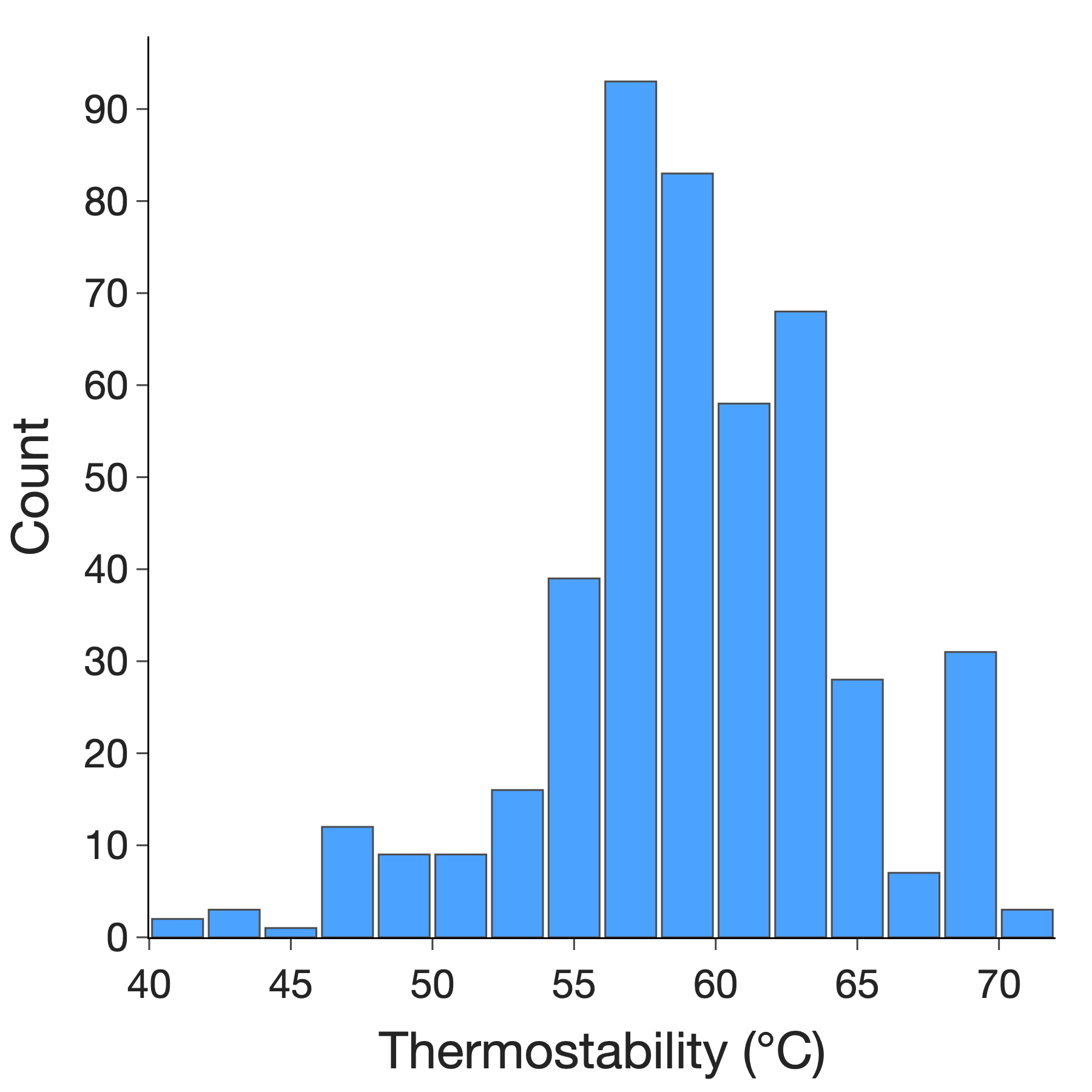}
  \caption{Anti-SARS-CoV-2 VHH }
\end{subfigure}
\hfill
\begin{subfigure}[t]{0.3\textwidth}
  \includegraphics[width=\linewidth]{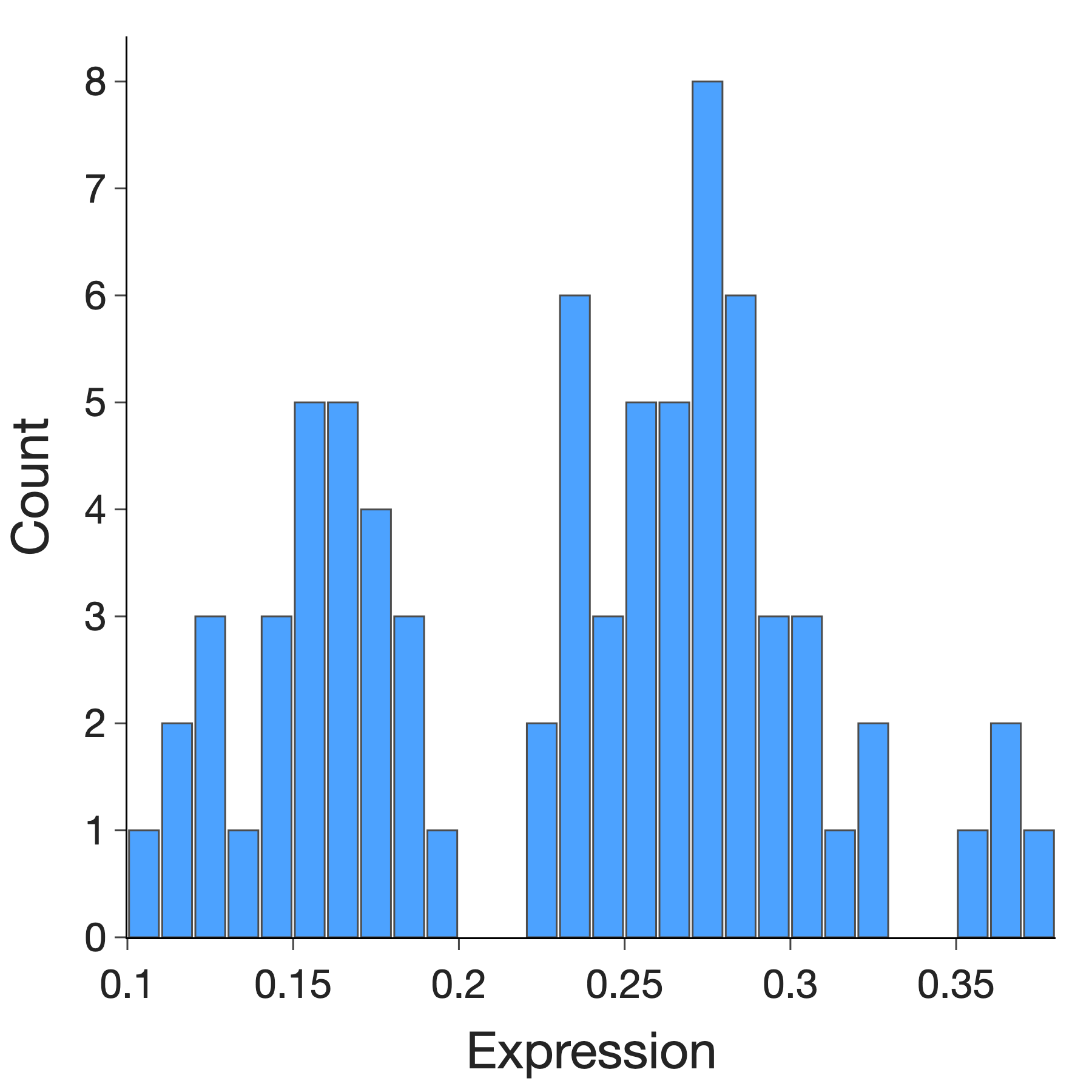}
  \caption{Trastuzumab scFv }
\end{subfigure}
\caption{Dataset distributions.}
\label{in-silico}
\end{figure}
\textbf{anti-SARS-CoV-2 VHH:} The wild type \cite{hanke2020alpaca} already expresses well and has a binding affinity in the single-digit nanomolar range.

\textbf{DhaA:} Haloalkane dehalogenase catalyzes the hydrolysis of halogenated compounds by cleavage of the carbon-halogen bond. 

\textbf{Trastuzumab scFv:} An scFv version of the cancer drug Trastuzumab, which expresses poorly and has a binding affinity of 3--5 nanomolar. 

\begin{table}[h]
\centering
\begin{tabular}{lcccc}
\toprule
\textbf{Protein} & \textbf{Function} & \textbf{$N$} & \textbf{Mutation breadth} & \textbf{WT measurement} \\
\midrule
anti-SARS-CoV-2 VHH & Thermostability & 462 & 47.1\% & 58.67 C \\
anti-SARS-CoV-2 VHH $L$ & Thermostability & 1833 & 92.4\% & 58.67 C \\
Trastuzumab scFv & Expression & 76 & 13.1\% & 0.28 \\
Haloalkane dehalogenase & Thermostability & 474 & 40.3\% & 45.09 C \\
\bottomrule
\end{tabular}
\vspace{1em}
\caption{Optimized function, dataset size ($N$), mutation breadth (\% of positions mutated), and wild type measurement for datasets used in experiments.}
\label{dataset-table}
\end{table}

\section{Additional results}

\subsection{\textit{In silico} results}
\begin{figure}[h]
\centering
\includegraphics[width=0.4\textwidth]{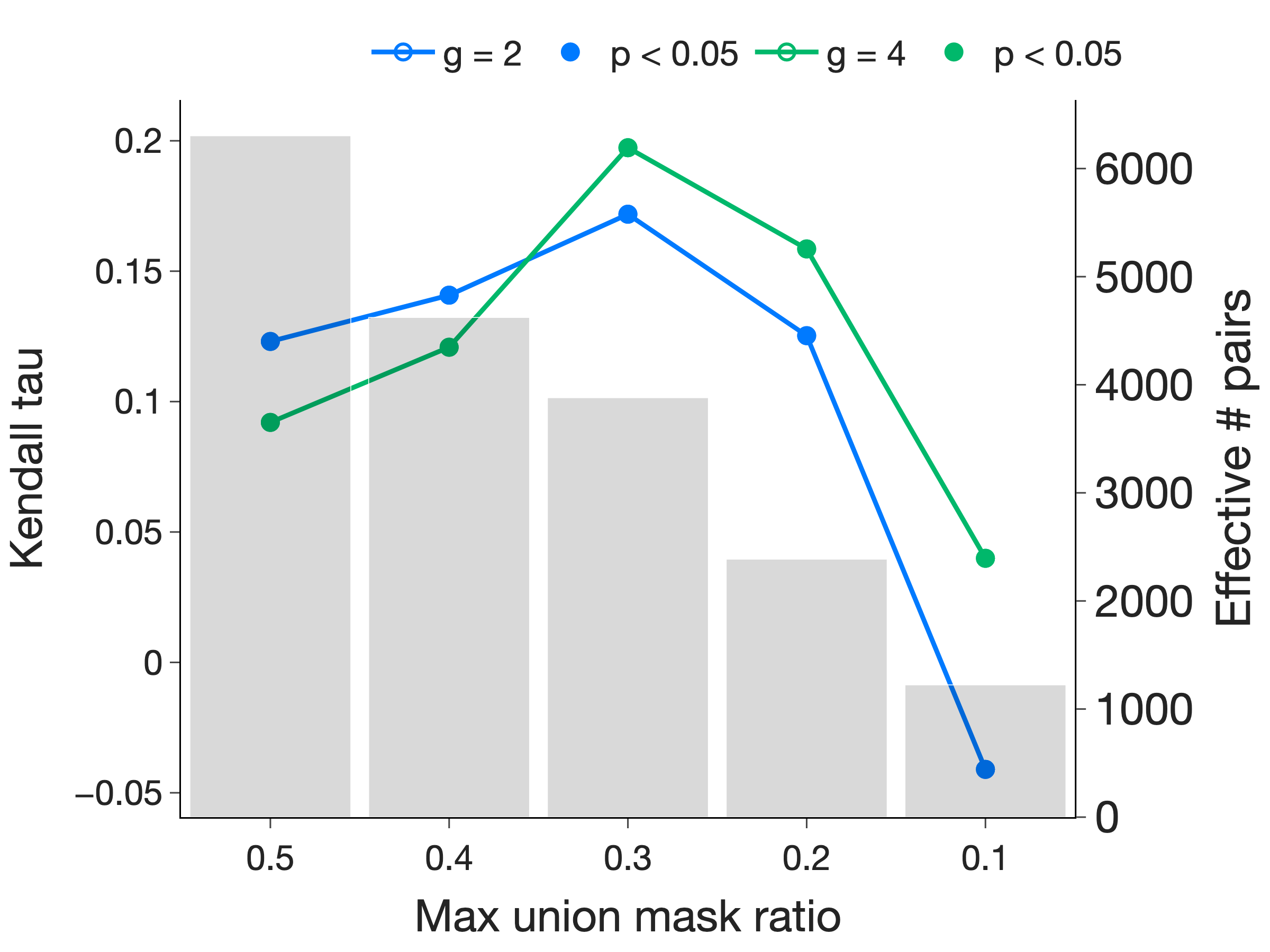}
\caption{Effect of clustering threshold $\tau$ on model performance and number of effective training pairs. Moderate clustering prunes redundant pairs without loss of performance.}
\label{kendall_all}
\end{figure}

We report Kendall’s $\tau$ as a complementary rank correlation metric to Spearman $\rho$. While Spearman captures monotonic correlation, Kendall’s $\tau$ measures the fraction of concordant versus discordant pairs and is therefore more sensitive to local ordering differences. Formally, it is defined as
\[
\tau = \frac{(\#\text{concordant pairs}) - (\#\text{discordant pairs})}{\#\text{total pairs}}.
\]
We report additional results with Kendall’s $\tau$ in Figure \ref{kendall_main} and Figure \ref{kendall_all}. The consistent trends across Spearman $\rho$ and Kendall's $\tau$ confirm that our conclusions are robust to the choice of ranking measure.

\begin{table}[h]
\centering
\vspace{1em}
\begin{tabular}{lcccccc}
\toprule
\textbf{Dataset} &
\multicolumn{2}{c}{\textbf{Training Time}} &
\multicolumn{2}{c}{\textbf{Total Run Time}} &
\multicolumn{2}{c}{\textbf{Speedup ($\times$)}} \\
\cmidrule(lr){2-3} \cmidrule(lr){4-5} \cmidrule(lr){6-7}
& g-DPO & DPO & g-DPO & DPO & Training & Total \\
\midrule
anti-SARS-CoV-2 VHH        & 1h49m21s & 3h03m44s & 1h52m58s & 3h07m35s & 1.68$\times$ & 1.66$\times$ \\
anti-SARS-CoV-2 VHH L      & 1h21m59s & 7h22m09s & 1h32m30s & 7h26m32s & 5.40$\times$ & 4.83$\times$ \\
Trastuzumab scFv           & 18m46s   & 1h00m29s & 22m06s   & 1h03m59s & 3.22$\times$ & 2.89$\times$ \\
Haloalkane dehalogenase    & 2h27m55s & 7h13m56s & 2h32m14s & 7h18m44s & 2.93$\times$ & 2.88$\times$ \\
\bottomrule
\end{tabular}
\vspace{1em}
\caption{Runtime comparison between g-DPO and DPO across datasets. Total run time includes training, clustering, and constant overhead. Training time is wall clock times for convergence. Speedup is computed as DPO / g-DPO.}
\label{tab:runtime_comparison}
\end{table}

\begin{figure}[h]
\centering
\includegraphics[width=0.4\textwidth]{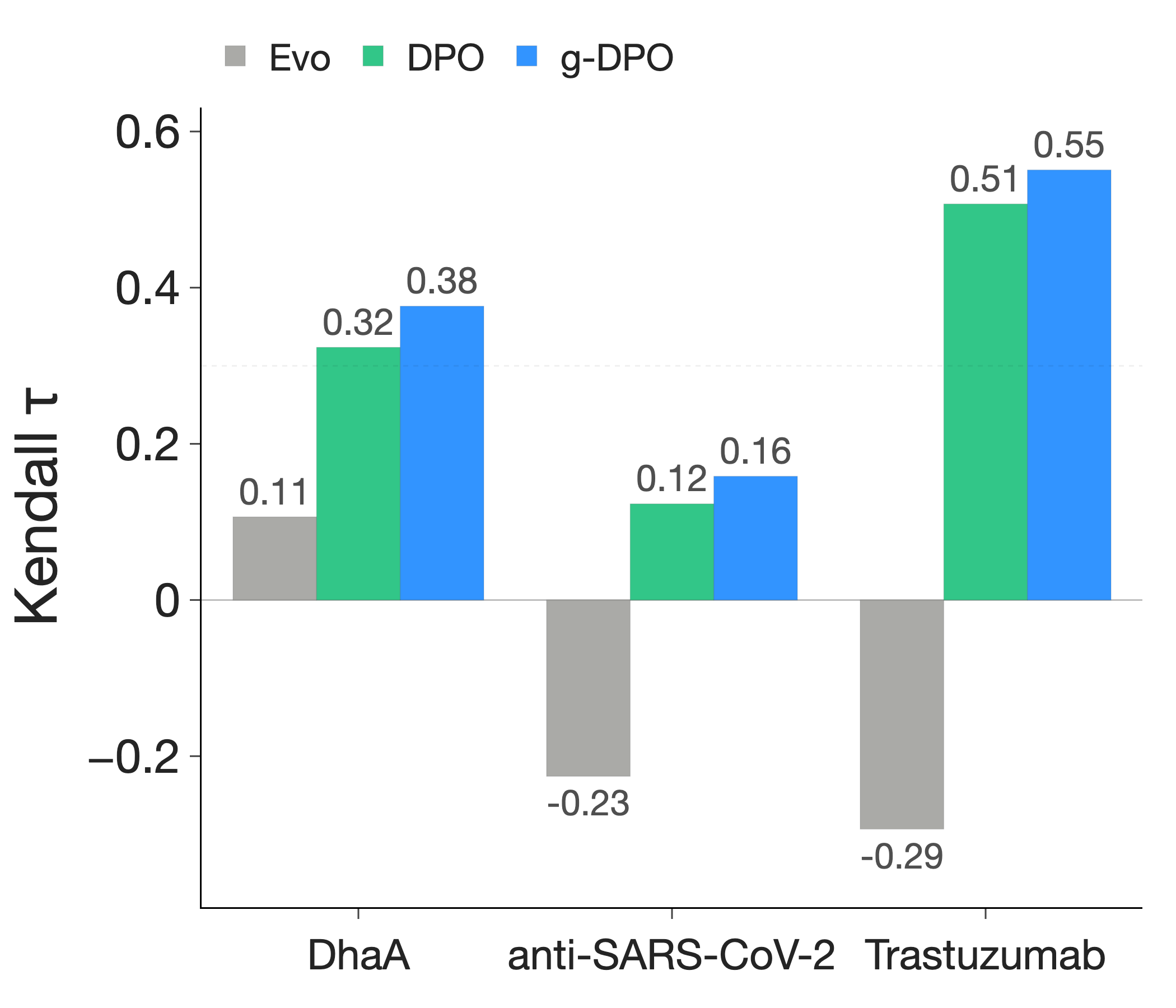}
\caption{Kendall tau ($\tau$) correlation between PLL scores and experimental measurements on holdout test sets for the evo-tune, DPO, and g-DPO models. }
\label{kendall_main}
\end{figure}

\subsection{Statistical significance tests}
We include two-sample Kolmogorov-Smirnov (KS) tests on the predicted property distributions of generated sequences (see Table \ref{ks-table}). While the KS tests indicate statistically significant differences between DPO and g-DPO, the effect sizes are very small, showing that the distributions are nearly identical in practice. This supports our conclusion that the computational gains of g-DPO do not come at the expense of model quality.

\begin{table}[h]
\centering
\begin{tabular}{lccccc}
\toprule
\textbf{Dataset} & \textbf{Pair} & \textbf{$D$} & \textbf{p-value} & \textbf{Statistic location} & \textbf{Sign} \\
\midrule
DhaA & g-DPO - DPO & 0.0290 & $<0.01$ & 51.90 & -1 \\
DhaA & g-DPO - Ref & 0.1736 & $<0.01$ & 52.49 & -1 \\
DhaA & DPO - Ref & 0.1851 & $<0.01$ & 52.07 & -1 \\
\midrule
Ty1 & g-DPO - DPO & 0.0385 & $<0.01$ & 56.99 & +1 \\
Ty1 & g-DPO - Ref & 0.5325 & $<0.01$ & 56.89 & -1 \\
Ty1 & DPO - Ref & 0.5695 & $<0.01$ & 56.90 & -1 \\
\midrule
Her2 & g-DPO - DPO & 0.0829 & $<0.01$ & 0.2506 & +1 \\
Her2 & g-DPO - Ref & 0.3477 & $<0.01$ & 0.2408 & -1 \\
Her2 & DPO - Ref & 0.3587 & $<0.01$ & 0.2411 & -1 \\
\bottomrule
\end{tabular}
\vspace{1em}
\caption{Two-sample Kolmogorov-Smirnov tests comparing predicted property distributions between models. 
The KS statistic $D$ reports the maximum distance between empirical CDFs; smaller values indicate closer distributions. 
$p$-values below $0.05$ indicate statistically significant differences. 
}
\label{ks-table}
\end{table}

\end{document}